\documentclass[lettersize,journal]{IEEEtran}
\usepackage{amsmath,amsfonts}
\usepackage{array}
\usepackage[caption=false,font=normalsize,labelfont=sf,textfont=sf]{subfig}
\usepackage{textcomp}
\usepackage{amsmath}
\usepackage{stfloats}
\usepackage{url}
\usepackage{verbatim}
\usepackage{graphicx}
\def\BibTeX{{\rm B\kern-.05em{\sc i\kern-.025em b}\kern-.08em
    T\kern-.1667em\lower.7ex\hbox{E}\kern-.125emX}}
\usepackage{balance}

\usepackage[numbers]{natbib}
\usepackage{multicol}
\usepackage{multirow}
\usepackage[bookmarks=true]{hyperref}
\usepackage{amsfonts}
\usepackage{amsmath} 
\usepackage{graphics} 
\usepackage{graphicx}
\usepackage{caption}
\usepackage{xcolor}
\usepackage{gensymb}

\usepackage{algorithm}
\usepackage{algpseudocode}
\usepackage{amssymb}
\usepackage{algorithmicx}

\usepackage{pgfplots}
\usepgfplotslibrary{groupplots}

\usepackage{booktabs}

\usepackage{pifont}
\newcommand{\cmark}{\ding{51}}%
\newcommand{\xmark}{\ding{55}}%

\usepackage[normalem]{ulem}
\useunder{\uline}{\ul}{}

\begin{document}
\title{I-CTRL: Imitation to Control Humanoid Robots Through Bounded Residual Reinforcement Learning}
\author{Yashuai Yan$^{*1}$, Esteve Valls Mascaro$^{*1}$, Tobias Egle$^{2}$, Dongheui Lee$^{1,3}$
\thanks{$^{1}$Yashuai Yan,  Esteve Valls Mascaro and Dongheui Lee are with Autonomous Systems Lab, Technische Universität Wien (TU Wien), Vienna, Austria (e-mail: \texttt{\{yashuai.yan, esteve.valls.mascaro, dongheui.lee\}@tuwien.ac.at}).\\$^{2}$Tobias Egle is with Robotics Systems Lab, Technische Universität Wien (TU Wien), Vienna, Austria (e-mail: \texttt{\{tobias.egle\}@tuwien.ac.at}). \\ $^{3}$Dongheui Lee is also with the Institute of Robotics and Mechatronics (DLR), German Aerospace Center, Wessling, Germany.}\\
\href{https://evm7.github.io/I-CTRL/}{\color{blue} evm7.github.io/I-CTRL}
}

\twocolumn[{%
\renewcommand\twocolumn[1][]{#1}%
\maketitle
\begin{center}
    \centering
    \vspace{-6mm}
    \captionsetup{type=figure}
    \includegraphics[width=.88\textwidth]{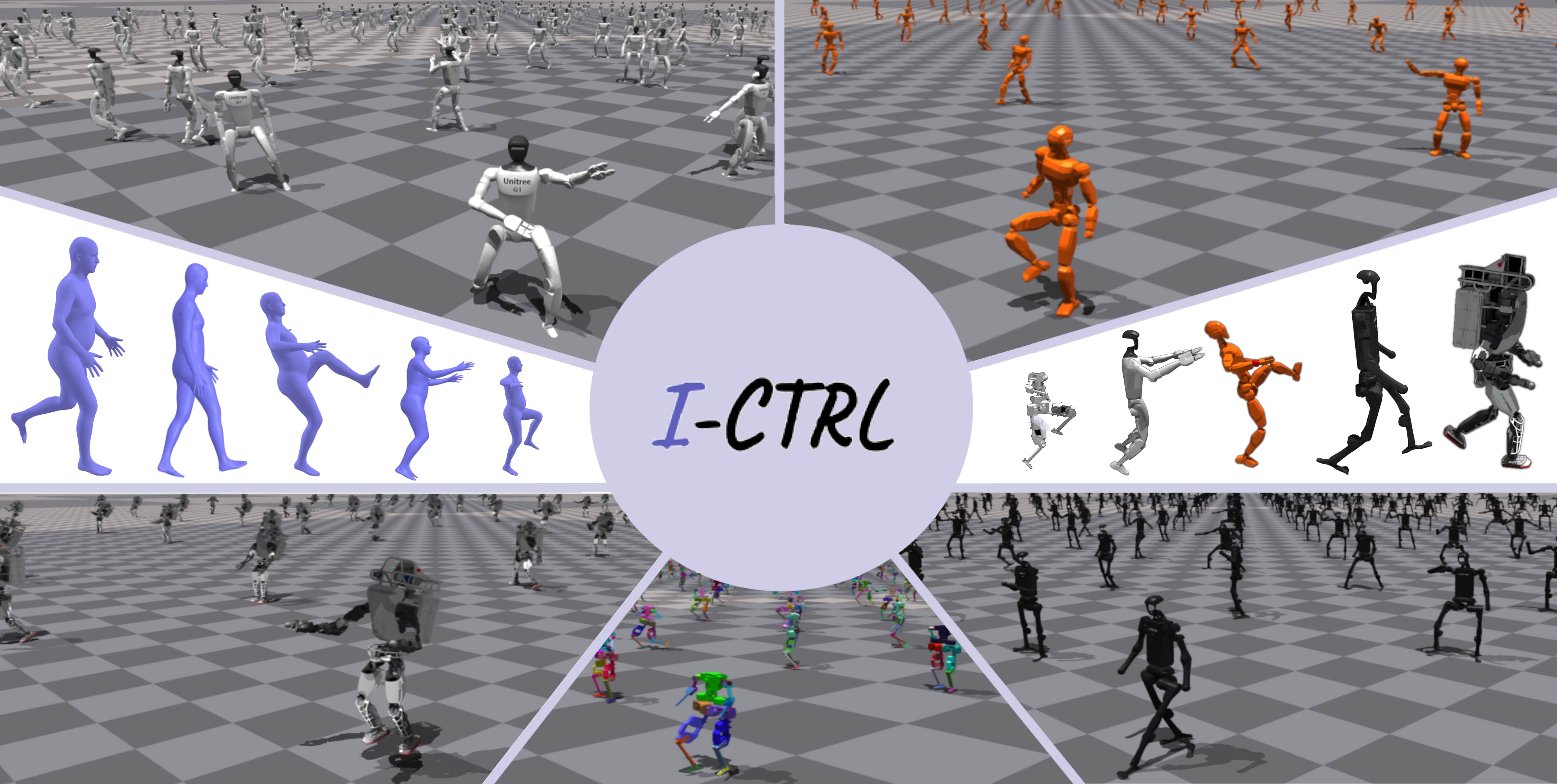}
    \captionof{figure}{Various bipedal humanoid robots imitate dynamic human motions through a reinforcement learning model, named I-CTRL. Each block shows diverse behaviors for the G1, JVRC-1, ATLAS, BRUCE, and H1 robots.}
    \label{fig:teaser}
\end{center}%
}]

\renewcommand{\thefootnote}{\fnsymbol{footnote}}
\footnotetext{$^{1}$Yashuai Yan,  Esteve Valls Mascaro, and Dongheui Lee are with Autonomous Systems Lab, Technische Universität Wien (TU Wien), Vienna, Austria (e-mail: \texttt{\{yashuai.yan, esteve.valls.mascaro, dongheui.lee\}@tuwien.ac.at}).}

\footnotetext{$^{2}$Tobias Egle is with Robotics Systems Lab, Technische Universität Wien (TU Wien), Vienna, Austria (e-mail: \texttt{\{tobias.egle\}@tuwien.ac.at})}

\footnotetext{$^{3}$Dongheui Lee is also with the Institute of Robotics and Mechatronics (DLR), German Aerospace Center, Wessling, Germany.}
\footnotetext{$^{*}$Contributed equally to the work.}

\markboth{Journal of IEEE Robotics \& Automation Magazine, Special Issue on Humanoid Robots, June~2024}%
{Y. Yan, E. Valls Mascaro, T. Egle, D. Lee, I-CTRL: Imitation to Control Humanoid Robots Through Constrained Reinforcement Learning}

\begin{abstract}
Humanoid robots have the potential to mimic human motions with high visual fidelity, yet translating these motions into practical, physical execution remains a significant challenge. Existing techniques in the graphics community often prioritize visual fidelity over physics-based feasibility, posing a significant challenge for deploying bipedal systems in practical applications. This paper addresses these issues through bounded residual reinforcement learning to produce physics-based high-quality motion imitation onto legged humanoid robots that enhance motion resemblance while successfully following the reference human trajectory. Our framework, Imitation to Control Humanoid Robots Through Bounded Residual Reinforcement Learning (I-CTRL), reformulates motion imitation as a constrained refinement over non-physics-based retargeted motions. I-CTRL excels in motion imitation with simple and unique rewards that generalize across five robots. Moreover, our framework introduces an automatic priority scheduler to manage large-scale motion datasets when efficiently training a unified RL policy across diverse motions. The proposed approach signifies a crucial step forward in advancing the control of bipedal robots, emphasizing the importance of aligning visual and physical realism for successful motion imitation.
\end{abstract}


\section{Introduction}
\IEEEPARstart{T}{he} task of endowing humanoid robots with human-like motion capabilities has been a longstanding challenge in robotics. A critical bottleneck in achieving this goal lies in imparting dynamic movements to robots that resemble the robustness and sophistication inherent in human motor skills. This paper addresses this challenge through a novel constraining technique for deep reinforcement learning that enables bipedal humanoid robots to efficiently learn various locomotion skills from humans, as shown in Figure \ref{fig:teaser}.

Recently, the animation and robotics communities have increasingly focused on translating human motions into various agents, each with a distinct focus. Motion imitation in animation has mainly addressed retargeting to preserve the visual resemblance and nuances of dynamics of human motions, while roboticists have focused on the physically plausible executions in real robots. Given the inherent difference in the tasks, two divergent approaches were taken.

With the emergence of large-scale motion capture (MoCap) datasets, the animation community adopted deep learning-based models to accurately translate human motions into simulated characters \cite{imitationnet}. However, those works in animation often do not satisfy the constraints existing in our physical world: foot penetrations, violations of joint limits and joint-velocity limits, or instability, among others. The task of ensuring physically plausible motions in the animated character has recently gained attention as an additional condition to ensure realism. To satisfy the plausibility of the motion, it is required to use a physics engine that emulates the real world and allows the algorithms to extend animation to reality. The success of Deep Reinforcement Learning (DRL) has motivated the animation community to use this technology to control physics-based animated characters \cite{deepmimic, amp, perceptual}. To generate more human-like dynamics, \citet{deepmimic, amp} reformulated the task as a goal-conditioned DRL problem, where human motion served as a reference to guide the DRL in learning. However, \citet{deepmimic} required manual tuning of the rewards for different motions, while \citet{amp} trained a discriminant network over a small-scale distribution of similar motions to act as a reward function, which greatly increased the complexity of training. Recently, \citet{perceptual} proposed the task of imitating large-scale motion datasets using a single policy. For that, \citet{perceptual} adopted a progressive RL with a mixture of experts (MoE) strategy. However, the aforementioned works \cite{deepmimic, amp, perceptual} rely on human data that match identically the kinematics of their simulated avatar, while not being challenged by different mass distributions, joint limits, and kinematic structures of robots.

To cope with these challenges, the robotics community has adopted control-based optimization algorithms \cite{lee,  motion_im_rl} for the whole-body control of bipedal robots. Historically, recurring motions such as bipedal walking and running have been addressed using biologically inspired template models, with particular emphasis on the linear inverted pendulum model (LIPM) \cite{invertedpendulum} for walking and the spring-loaded inverted pendulum \cite{6697099} for running. Later, model predictive control (MPC) \cite{mpc_1} showed improvements in generating stable center-of-mass (COM) trajectories based on future footsteps. The current state-of-the-art in robot control adopts divergent component of motion (DCM) trajectory generation \cite{dcm} with MPC-based step adaptation \cite{9082021} for the locomotion of full-scale humanoid robots. However, those frameworks require high configuration tuning and focus only on simple walking behaviors, without allowing for versatile human motion imitation. Lately, MPC-based approaches have been combined with deep learning to produce robot motions that resemble humans or animals from casual videos \cite{motion_im_rl}. However, \citet{motion_im_rl} required eight hours and eight GPUs to imitate a single one-minute video, making it infeasible for real-time imitation and generalization to large-scale motions. On the other hand, DRL has already shown promising potential in producing agile and robust locomotion for quadrupedal robots \cite{agarwal2023legged} which can work in real time and be deployed in unconfined outdoor environments. In terms of bipedal locomotion, \citet{rl_bipedal2} adopted Deep RL to produce specific motions: walk, run, or jump. Xie et al. \cite{xie2018feedbackcontrolcassiedeep} proposed learning bipedal walking policies by learning residual joint angles on a reference motion rather than predicting absolve joint angles. Smith et al. \cite{smith2022walkparklearningwalk} limited the RL action space to a narrow range centered around constant default joint angles to improve the data efficiency. Recently, Cheng et al. \cite{cheng2024express} extended DRL to learn a single policy for large-scale motion diversity in an H1 bipedal robot but focused solely on the imitation of the upper body, facilitating balance without restricting the legs. To enhance whole-body imitation, He et al. \cite{he2024learning} adapted the mixture of expert models from \cite{perceptual} to the H1 robot and showcased sim-to-real experiments. However, Luo et al. \cite{perceptual} required large computational resources for training, several ensemble models, and did not achieve high-quality style resemblance. Similarly, Fu et al. \cite{fu2024humanplus} fine-tuned a DRL policy model with visual feedback to perform specific tasks autonomously from 50 demonstrations.

In this work, we demonstrate that bipedal humanoid robots can learn massive motion skills efficiently by imitating humans through our I-CTRL. Contrary to previous works that require massive training and specific reward tuning for each robot, we constrain the exploration region in reinforcement learning to effectively address motion imitation tasks in a general manner, showcasing the robustness of I-CTRL across five different robots. To achieve this, we adopt ImitationNet \cite{imitationnet} to derive reference motions from an extensive human dataset. However, while ImitationNet excels at preserving the motion style, it ignores the dynamics of the retargeted motions, which can result in failures when applied in real-world scenarios. Therefore, I-CTRL refines the reference motions from ImitationNet to ensure feasible robot motion in the physical world and scalability to a large diversity of motions. Additionally, we propose a general training policy that enables the control of five bipedal humanoid robots without tuning the reward signals: BRUCE, Unitree-H1, Unitree-G1, ATLAS, and JVRC-1. The versatility of I-CTRL is crucial to improving rapid adaptation to new robots. Finally, to overcome the highly constrained and complex nature of the different kinematics while ensuring efficient training and preserving the motion style, we proposed a novel method to reduce the exploration of the DRL agent while training. Finally, we designed a self-adapted curriculum learning strategy that continuously encourages the model to learn challenging motions during training, resulting in a boost in speed and performance. This results in a single policy that can accurately follow around 9K different dynamic motions only with 15 hours of training. 
Our efforts resulted in the following contributions.

\begin{enumerate}
    \item Designing a bounded residual reinforcement learning framework that effectively preserves the style of the targeted motions and efficiently generalizes to around 9K motions using a single policy and shared rewards.
    
    \item Proposing a motion selection strategy that automatically prioritizes large-scale human motions, allowing for the training of a single policy on extensive motion datasets within just 15 hours.
    
    \item Learning physics-based human-like motions for various bipedal humanoid robots.
    
    \item Evaluating I-CTRL quantitatively and qualitatively on five bipedal robots and a variety of motions to showcase the versatility of our model without reward tuning.

\end{enumerate}

\section{Methodology}
\begin{figure*}[]
    \centering
    \includegraphics[width=0.99\textwidth]{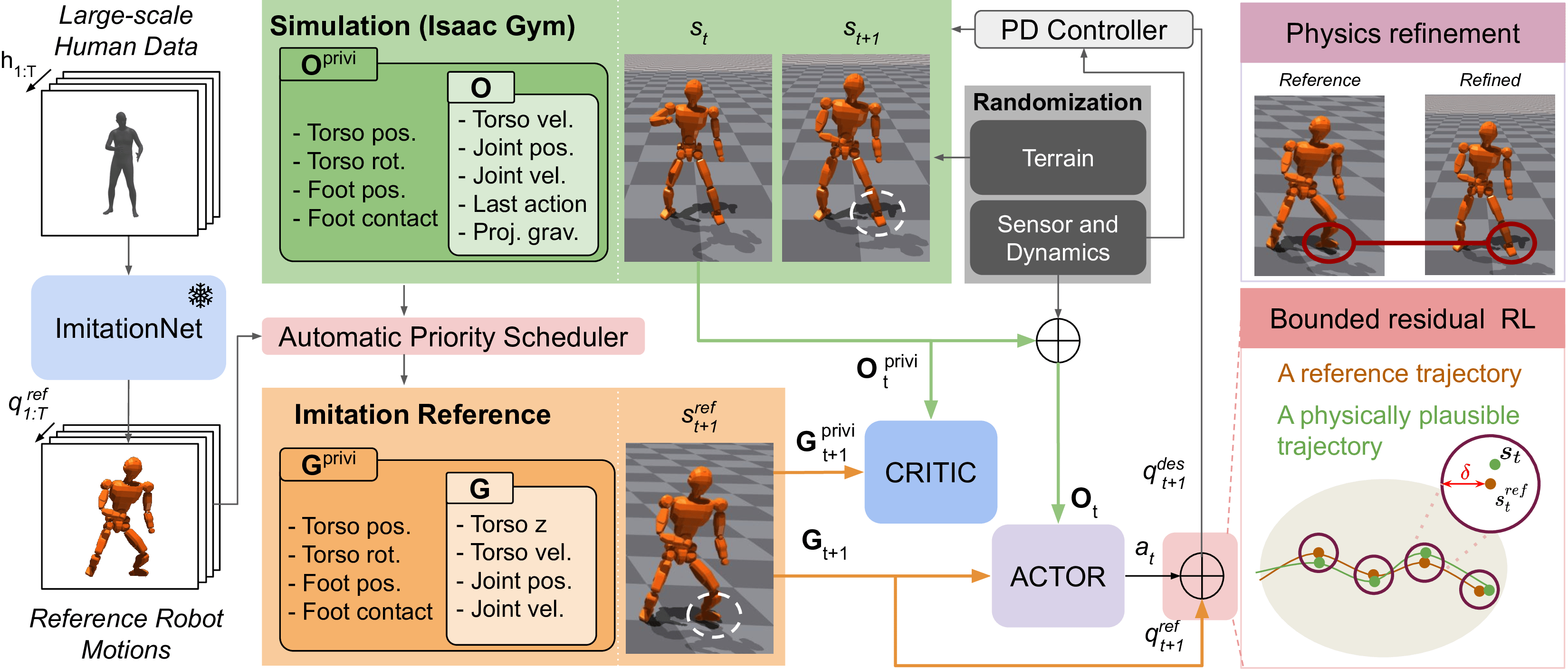}
    \caption{\textbf{Architecture overview of our I-CTRL framework.} First, we train ImitationNet \cite{imitationnet} to retarget human motions into different robots. ImitationNet provides stylistic robot poses $\mathbf{q}^{ref}_{1:T}$ that resemble the original human but are not feasible for executing in the physical world.  We then use our I-CTRL framework to train a whole-body controller that refines the reference motions in a physics simulator. I-CTRL introduces additional constraints to a standard residual reinforcement learning process. These constraints ensure the physical plausibility of various human motions, while significantly and effectively reducing the exploration space for reinforcement learning, allowing it to train a control policy across thousands of dynamic human movements within hours.}
    \label{fig:modeloverview}
\end{figure*}

\noindent In this section, we present our whole-body imitation algorithm for humanoid robots. An overview of our framework is illustrated in Fig. \ref{fig:modeloverview}.

\subsection{Problem formulation}
\label{sec: problem_formulation}
Let $\mathbf{h}_t \in \mathbb{R}^{J\times n}$ be a human pose at time $t$ composed of $J$ joints in local quaternion representation ($n=4$), and $\mathbf{H}=[\mathbf{h}_{1}, \cdots, \mathbf{h}_{T}]  \in \mathbb{R}^{T \times J \times n}$ a human motion. Similarly, $\mathbf{q}_t \in \mathbb{R}^{J\times 1}$ represents a robot pose described in joint angles. As a notation convention, we use $\mathbf{q}^{ref}_t$ and $\mathbf{q}^{ref}_{1:T}$ to denote the robot reference pose and trajectory, respectively. For goal-conditioned RL, robot observation $\mathbf{O}_t$ consists of only their proprioception: joint angles $\mathbf{q}_{t}$, joint velocities $\dot{\mathbf{q}}_{t}$, linear and angular velocity in $\mathbf{v}_{t}, \boldsymbol{\omega}_{t} \in \mathbb{R}^{3}$, the last action $\mathbf{a}_{t}$ and projected gravity. Similarly, we consider $\mathbf{q}_{t}^{\text{ref}}$, $\dot{\mathbf{q}}_{t}^{\text{ref}}$, $\mathbf{v}_{t}^{\text{ref}}, \boldsymbol{\omega}_{t}^{\text{ref}}$, and additional torso height as the goal $\mathbf{G}_t$. Instead of assuming a symmetric actor-critic, where the actor and critic networks assess the same information, we extend the observation $\mathbf{O}$ and the goal $\mathbf{G}$ with privileged information (i.e., global position and orientation, the local feet movement, and their contact) to form $\mathbf{O}^{\text{privi}}$ and $\mathbf{G}^{\text{privi}}$ specifically for the critic network. Overall, our physics-based motion imitation task consists of first translating human motions $\mathbf{H}$ into robot reference motions $\mathbf{q}^{ref}_{1:T}$ and then refining these reference motions into robot motions $\mathbf{q}_{1:T}$ that satisfy the physical laws of the real world.

\subsection{Human-to-robot pose retargeting}
\label{sec:retarget}
Our task initially requires a retargeting process from human $\mathbf{h}_{t}$ to a robot pose $\mathbf{q}^{ref}_{t}$ that we use as a reference. To achieve $\mathbf{q}^{ref}_{t}$ so that it visually resembles the original human pose $\mathbf{h}_{t}$, we use ImitationNet \cite{imitationnet}. We adopt ImitationNet among the human-to-robot retargeting models, as it uses unsupervised learning in the retargeting process, which aligns with our purpose of building a system that can generalize over various robots with few modifications. To this end, we train a model $f$ based on ImitationNet for each robot, and the model is applied to map a human pose $\mathbf{h}_{t}$ to a robot pose $\mathbf{q}^{ref}_{t}$ ($f(\mathbf{h}_{t})=\mathbf{q}^{ref}_{t}$).

\subsection{Bounded residual reinforcement learning} 
We address physics-based motion retargeting as a process of constraint refinement applied to reference motions. To achieve this, we leverage residual RL to enhance the reference joint trajectories, similar to \cite{xie2018feedbackcontrolcassiedeep}. Furthermore, we impose constraints on these augmentations to ensure that the retargeted motions maintain visual fidelity. Unlike \cite{smith2022walkparklearningwalk}, which applies constant default limits to the joint space, our method confines RL exploration to a dynamic hypertube centered on the reference trajectories (see Fig. \ref{fig:modeloverview}).

Formally, we formulate our I-CTRL framework as a constraint Markov Decision Process (CMDP), defined as $\left \langle  \mathcal{S, A}, T, R, \mathcal{C}, \rho, \gamma  \right \rangle$. Unlike standard MDPs, a CMDP includes a set of constraint functions in $\mathcal{C}$. Accordingly, we define the objective and constraint functions in I-CTRL as follows:

\begin{equation}
\begin{gathered}
 \underset{x}{\text{maximize}} \;\;\; \mathop{\mathbb{E}}_{\rho, \pi, T} \left[ \sum_{t=0}^{\infty} \gamma^{t} R(s_{t}, a_{t}, s_{t+1}) \right] \\
 \text{s. t.}\;\;\;\;  D(T(s_{t}, \pi(s_{t})), f(h_{t+1})) < \delta   
\end{gathered}
\label{eq:opt}
\end{equation}
where $D(\cdot, \cdot)$ indicates the distance metric between two states, $\pi$ is the policy to optimize, and $f$ is any function that maps human motions into state space $\mathcal{S}$. As discussed in Sec. \ref{sec:retarget}, ImitationNet \cite{imitationnet} is employed as $f$ in this work due to the high resemblance of retargeted motions. The exploration region in RL for searching optimal solutions is bounded by $\delta$, which serves as an upper limit on the allowable deviation between robot and human motion.
The robot states and the transition function $T$ are determined by the real-world environment or a physics simulator while our policy $\pi$ computes per-step action $\mathbf{a}_{t} \in \mathcal{A}$ conditioned on the state $\mathbf{s}_{t}$ and the reference $\mathbf{s}_{t}^{ref}$. The action $\mathbf{a}_{t}$ specifies target joint positions for PD controllers at each joint. The reward function $R$ computing per-step reward $r_{t} = R(\mathbf{s}_{t}, \mathbf{a}_{t}, \mathbf{s}_{t+1})$ is defined in Sec.\ref{sec:reward}. Similar to prior works \cite{deepmimic, amp, perceptual}, the proximal policy optimization (PPO) \cite{schulman2017proximal} is used to learn our imitation policy $\pi$.

\subsection{Rewards design}
\label{sec:reward}

\begin{table}[]
\centering
\resizebox{0.98\linewidth}{!}{%
\renewcommand{\arraystretch}{1.7}

\begin{tabular}{l|c|c|cc}
\textbf{Reward} & \textbf{Expression} & \textbf{Distance} &  $\boldsymbol{\omega}$ & $\boldsymbol{\lambda}$ \\ \midrule
$R^{\text{torso\_pos}}$ & \multirow{6}{*}{$\omega e^{-\lambda D}$} & $||\mathbf{p}^{\text{torso}} - \mathbf{p}_{\text{ref}}^{\text{torso}}||_{2}$ & 0.5 & 10 \\
$R^{\text{torso\_rot}}$ & & $||\boldsymbol{\theta}^{\text{torso}} \ominus \boldsymbol{\theta}_{\text{ref}}^{\text{torso}}||_{2}$ & 0.5 & 10 \\
$R^{\text{joint}}$ & & $||\mathbf{q} - \mathbf{q}_{\text{ref}}||_{2}$ & 1 & 5 \\
$R^{\text{foot}}$ & & $||\mathbf{p}^{\text{foot}} -\mathbf{p}_{\text{ref}}^{\text{foot}}||_{2}$ & 0.5 & 10 \\
$R^{\text{energy}}$ & & $||\boldsymbol{\tau} \cdot  \dot{\mathbf{q}}||$ & $10^{-6} $ & 1.0 \\
$R^{\text{smooth}}$ &   & $|| \ddot{\mathbf{q}}||_{2}$ & 0.3  & 0.1 \\ \midrule
$R^{\text{contact}}$ & \multirow{2}{*}{$\left\{\begin{matrix} -\lambda & \text{Condition}\\ 0 & \text{otherwise} \end{matrix}\right.$}  & foot contact wrong  & - & 1 \\
$R^{\text{terminate}}$ &                   & robot falls & - & 20
\end{tabular}
}
\caption{System rewards}
\label{table:rew}
\end{table}

The reward functions in this work are illustrated in Table \ref{table:rew}. One important goal of our I-CTRL is to alleviate the burden of designing tens of reward functions as in previous works \cite{cheng2024express, he2024learning}. We apply the same reward functions and parameters across all robots to showcase the generalization of I-CTRL.  The rewards in Table \ref{table:rew} include task-specific rewards ($R^{\text{torso\_pos}}$ and $R^{\text{torso\_rot}}$) aimed at tracking the human's position and rotation, as well as style rewards ($R^{\text{joint}}$ and $R^{\text{foot}}$) to replicate the reference motion's style. Furthermore, we penalize energy consumption with $R^{\text{energy}}$ and minimize motion jitter through $R^{\text{smooth}}$. Two more terms, $R^{\text{contact}}$ and $R^{\text{terminate}}$, are introduced to promote stable contact with the ground and body balance during retargeting.

\subsection{Domain randomization}

\begin{table}[]

\centering
\resizebox{0.85\linewidth}{!}{%
\begin{tabular}{lccc}
Parameter & Unit & Distribution & Operator \\ \midrule
\multicolumn{4}{c}{\textbf{Sensory Noise}}  \\ \midrule
Angular Velocity & rad/s  & $\mathcal{U}(-0.1, 0.1)$ & additive \\
Joint Position & rad  & $\mathcal{U}(-0.1, 0.1)$ & additive \\
Joint Velocity & rad/s & $\mathcal{U}(-1.0, 1.0)$ & additive \\ \midrule
\multicolumn{4}{c}{\textbf{Dynamics Noise}}  \\ \midrule
PD Factor & - & $\mathcal{U}(0.75, 1.25)$ & scaling \\
Link Mass & Kg & $\mathcal{U}(0.7, 1.3)$ & scaling \\
Friction & - & $\mathcal{U}(0.5, 2.0)$ & scaling  \\ \midrule

\multicolumn{4}{c}{\textbf{Environmental Noise}}  \\ \midrule
\multicolumn{4}{c}{\text{flat terrain}, \text{rough terrain}} 
\end{tabular}
}
\caption{Domain randomization in simulation.}
\label{table:domain_randomization}
\end{table}

Domain randomization (DR) has proven to be essential to overcome the sim-to-real gap \cite{he2024learning, fu2024humanplus}. Our approach focuses on capturing real-world variability from three key sources: sensors, dynamics, and environment. Sensory variability is introduced by adding noise to observations, such as angular velocity and joint positions and velocities. Motor variations are addressed by adjusting the motor parameters, including stiffness and damping in the controller settings. We also randomize the link masses to account for potential discrepancies between real robots and the URDF-defined models. Finally, we also randomize environmental friction and train policies on rough terrain to better handle disturbances encountered in real-world settings. The detailed parameters are shown in Table \ref{table:domain_randomization}.

\subsection{Automatic priority scheduler}
\label{sec:aps}
\begin{algorithm}

\caption{Automatic Priority Scheduler}\label{algo:aps}
\renewcommand{\algorithmicindent}{0.5em} 
\begin{algorithmic}[1]

\State \textbf{Initialization:}

\State $N_j, P_j \gets 0, 0 \; \forall \; M_j$ \Comment{Completion count and priority}
\State $S_j \gets \text{random}(0, \text{len}(M_j))\; \forall \; M_j$ \Comment{Random start index}
\State $C_j \gets M_j[S_j : \text{len}(M_j)] \; \forall \; M_j$  \Comment{Clip starting at $S_i$}
\State
\State $i  \gets \text{sample by priority}(P)$

\While{training is not complete}
    \If{check\_success($C_i$)} \Comment{Reaches end clip}
        \If{$S_i == 0$} \Comment{Motion completed}
            \State $N_i \gets N_i + 1$ \Comment{Increment success count}
            \State $P_i \gets 1 - \frac{N_i}{\max(N)}$ \Comment{Update priority}
            \State $i  \gets \text{sample by priority}(P)$
        \Else 
            \State $S_i \gets \text{random}(0, S_i)$ \Comment{Increase clip length}
            \State $C_i \gets M_i[S_i : \text{len}(M_i)]$ \Comment{Update clip}

        \EndIf
    \EndIf
\EndWhile
\end{algorithmic}
\end{algorithm}

We introduce the Automatic Priority Scheduler (APS) to train a unified policy on a large-scale human motion dataset. Initially, APS assigns equal priority to all motions, starting each at a random point in its sequence. During training, if the policy successfully handles a motion clip, its length increases. Upon completion of the entire motion, its success count is increased, and its priority is then recalculated accordingly. After completing one motion, the system resamples another motion for training based on the updated priorities. The details of APS are shown in Algorithm \ref{algo:aps}. APS differs from typical curriculum learning strategies, where the user has to define the complexity of each motion, and from the progressive RL \cite{perceptual, he2024learning} that offline schedules motion subsets, leading to overfitting of different expert modules.

\section{Experiments}
\subsection{Human MoCap dataset} 
We select 8.855 daily motions, including \emph{walking},   \emph{cooking},  \emph{punching},  \emph{dancing},  \emph{jumping}, 
\emph{running in a circle}, \emph{strong turns while dancing}, etc., from the HumanML3D dataset which contains 14616 high-quality human motions captured through motion capture systems (MoCaP). We excluded motions that are unfeasible in physics-based simulators without the necessary environmental data, such as \emph{swimming} or \emph{walking up stairs}. In addition, we excluded highly dynamic motions that are difficult to perform even for ordinary humans, such as \emph{parkour},  \emph{break-dance}, or \emph{handstand}. 

\subsection{Humanoid robots} 
To demonstrate the generalization capabilities of our algorithm, we evaluate our motion imitation system on five humanoid robots of varying kinematics and dynamics. Table \ref{tab:robots} presents the properties of various robots, with masses ranging from 4.7 to 181.0 kilograms and heights ranging from 70 to 183 centimeters. As the sizes of different robots can vary a lot from the human model, we normalize the human root trajectories by multiplying a scale factor on the root's position and linear velocity. The scaling factor is computed as the ratio between the robot's height and the human's height and is indicated in Table \ref{tab:robots}.

\begin{table}[]
\centering
\resizebox{0.95\linewidth}{!}{%
\begin{tabular}{l|ccccc}
 & ATLAS & H1 & JVRC-1 & G1 & BRUCE \\ \midrule
DoF & 23 & 19 & 23 & 21 & 16 \\
height (cm)  & 183 & 180 & 140 & 132 & 70 \\
weight (kg) & 181.0 & 47.0 & 62.2  & 35.0  & 4.8 \\
human to robot (\%) & 1 & 1 & 0.8 & 0.75 & 0.45
\end{tabular}
}
\caption{Properties of different robots. Note that the human-to-robot is calculated as the leg-length ratio between the human character used in \cite{luo2021dynamics, perceptual} and each robot.}
\label{tab:robots}
\end{table}

\subsection{Experimental setting} 
For reinforcement learning, both actor and critic are implemented as neural networks with eight fully connected linear layers. Each linear layer has 256 neurons. The constraint in our optimization problem (see Equation \ref{eq:opt}) is implemented in the robots' joint space by limiting the deviation $\mathbf{q}^{des}_t = \mathbf{q}^{ref}_t + \mathbf{a}_t$ to a maximum of $\delta$. We set $\delta_{arms}=10$ $\degree$ for the arm joints to ensure high stylistic preservation of the reference motions, but we relax the constraints in the legs and torso $\delta_{\{legs, torso\}}=30$ $\degree$ to ensure sufficient exploration to maintain physical plausibility over diverse motions. These values are used for all robots in our experiments, although we acknowledge that fine-tuning them for each specific robot could enhance the results. We utilize NVIDIA's IsaacGym physics simulator to parallelize our training on an NVIDIA A4000 GPU. During training, the control policy is running at 60 Hz, while the simulation runs at 120 Hz. The experimental settings are shared among robots.

\subsection{Metrics} We evaluate I-CTRL on human daily motions from the HumanML3D dataset with the following metrics:
\begin{itemize}
    \item \textbf{Living Rate} indicates whether a robot has fallen at any point during the motion sequence.

    \item \textbf{Success Rate} evaluates if our robots are following the reference motions successfully. We adapt this metric from \citep{luo2021dynamics, perceptual}, where the human character is considered to fail when their root position deviates more than 0.5 meters from the reference root position. In our case, we scaled this distance based on the corresponding human-to-robot ratio, indicated in Table \ref{tab:robots} (e.g, 0.4 meters for JVRC-1).
    
    \item \textbf{Style Distance} assesses the visual similarity of imitated motions to reference ones. Style distance is defined as the Mean Squared Error (MSE) of the $xyz$-euclidean positions of joints in meter units.

    \item \textbf{Root Position Distance} showcases the ability to follow the root trajectories in a global coordinate system. The Root Position Distance is also considered as the MSE of the root position between the reference and the robot motion.
    
    \item \textbf{Root Rotation Distance} evaluates the ability of the robot to follow the orientation of human motion in a global coordinate system. The rotation distance is computed as the quaternion distance between the orientation of the robot's root and the reference motion.
    
\end{itemize}

\subsection{Evaluation} 
\subsubsection{Ablation Study} We evaluate the impact of our training strategies for the G1 robot in Table \ref{tab:ablationstudy}. The results demonstrate that incorporating our constraining technique (CT), privileged information (Priv), and APS-based training leads to significant improvements in learning a unified policy for large-scale motions, resulting in a $+10\%$ increase in success rate. Additionally, we assess the performance of policies trained with different robot state histories under domain randomization. Our findings indicate that feeding only the most recent robot state enhances the model’s generalization.

\begin{table*}[]
\centering

\resizebox{0.98\textwidth}{!}{%
\begin{tabular}{ccccccccc}
\textbf{History} & \textbf{CT} & \textbf{Priv.} & \textbf{APS} & \textbf{Living Rate (\%)} $\uparrow$ & \textbf{Success Rate (\%)} $\uparrow$ & \textbf{Root Pos. Dist. (m)} $\downarrow$ & \textbf{Root Rot. Dist. (\degree)} $\downarrow$ & \textbf{Style Dist. (m) $\downarrow$} \\ \midrule
 \multicolumn{9}{c}{\textbf{w/o Domain Randomization}}  \\ \midrule
 1 & \xmark & \xmark & \xmark & 93.08 & 78.28 & 0.162 & 0.142 & 0.100 \\
1 & \cmark & \xmark & \xmark & 95.81 & 83.90 & 0.128 & 0.127 & 0.020 \\
 1 & \cmark & \cmark & \xmark & 94.91 & 86.96 & 0.103 & {\ul 0.098} & 0.020 \\
1 & \cmark & \xmark & \cmark & \textbf{98.00} & {\ul 88.63} & {\ul 0.099} & 0.117 & \textbf{0.019} \\
1 & \cmark & \cmark & \cmark & {\ul 97.99} & \textbf{94.16} & \textbf{0.082} & \textbf{0.094} & {\ul 0.020} \\ \midrule
 \multicolumn{9}{c}{\textbf{w/ Domain Randomization}}  \\ \midrule
 5 & \cmark & \cmark & \cmark & 97.37 & 92.46 & 0.122 & 0.164 & 0.028 \\
1 & \cmark & \cmark & \cmark & \textbf{97.83} & \textbf{94.08} & \textbf{0.088} & \textbf{0.102} & \textbf{0.025} 
\end{tabular}
}
\caption{\textbf{Ablation study of the different training strategies for training a single policy for large-scale motions in the G1 robot}. Here, `CT' indicates the use of our proposed joint-space constraints, `Priv.' stands for privileged information in the critic, and `APS' for the use of the Automatic Priority Scheduler.}
\label{tab:ablationstudy}
\end{table*}

\subsubsection{Generalization to various robots} 
\begin{figure*}[]
    \centering
    \includegraphics[width=0.98\textwidth]{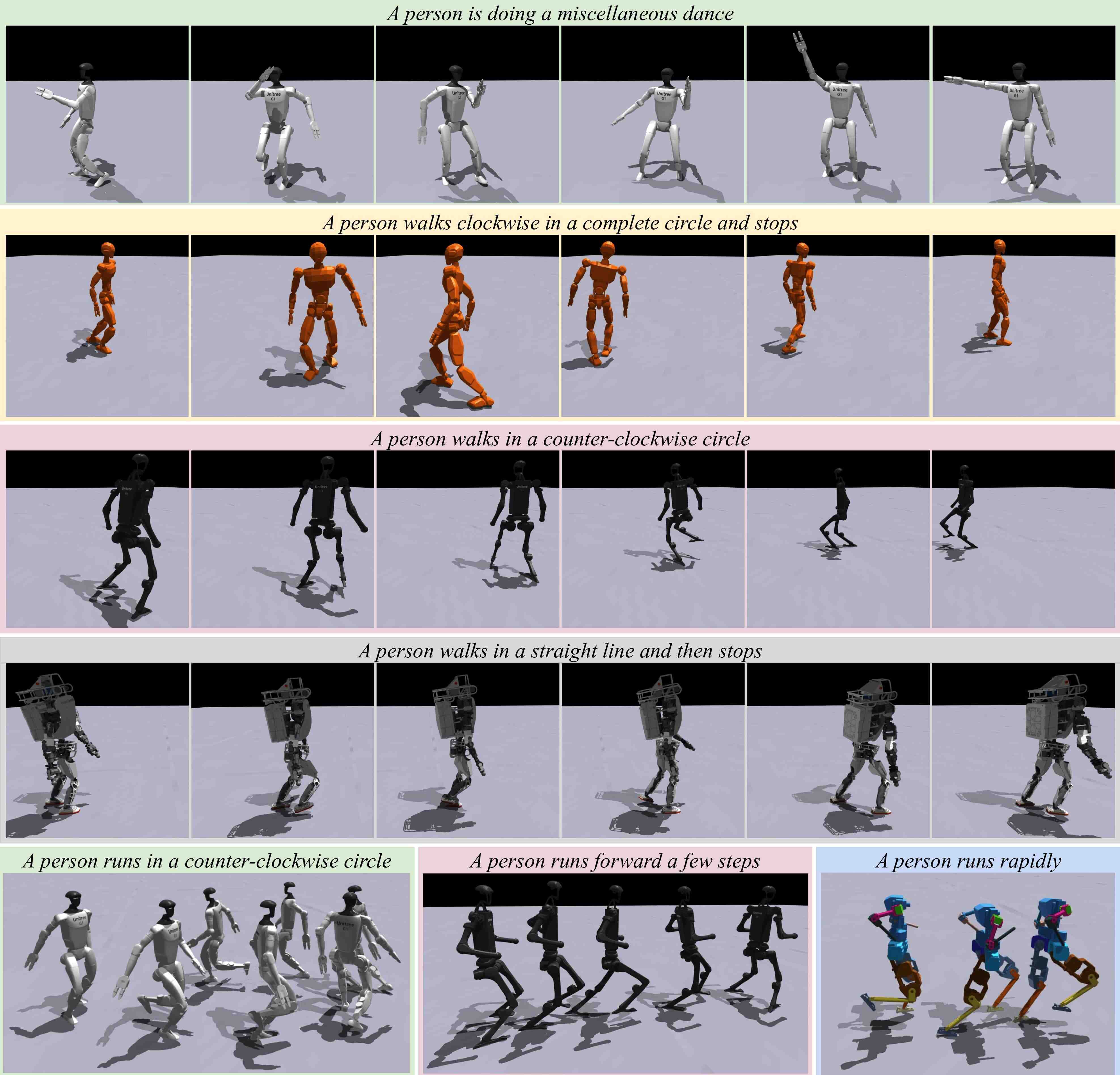}
    \caption{\textbf{Control of different bipedal humanoid robots for various motions} We showcase the results of our proposed method, I-CTRL, imitating various dynamic human motions described by the given text. The simulation is conducted with domain randomization and on uneven terrains.}
    \label{fig:text2motion}
\end{figure*}

We adopt the best configuration from the ablation study (Table \ref{tab:ablationstudy}) and accordingly train a policy for each of the five proposed robots. Note that we do not perform any reward or parameter tuning for a specific robot. We present these results in Table \ref{tab:universal_quantiative}, which shows that our I-CTRL can achieve a success rate of 82.94\% of the 8.855 motions on average in all robots. Note that ATLAS does not perform as well as other robots, primarily because it is significantly heavier. In our dynamic model, the randomized link mass can fluctuate by up to 50 kg, which impacts its performance. Fig. \ref{fig:text2motion} illustrates the qualitative results of our RL agent on some of these diverse motions.

\begin{table*}
\centering

\resizebox{0.98\linewidth}{!}{%
\begin{tabular}{@{}l|ccccc@{}}
Robots & Living Rate (\%) $\uparrow$ & Success Rate (\%) $\uparrow$ & Style Distance (m) $\downarrow$ & Root Pos. Dist. (m) $\downarrow$ & Root Rot. Dist. (\degree) $\downarrow$ \\ \midrule
H1 & 86.75 & 80.03 & 0.084 & 0.158 & 0.142 \\ 
BRUCE & 94.45 & 80.36 & 0.018 & 0.089 & 0.136 \\
ATLAS & 85.56 & 74.48 & 0.189 & 0.176 & 0.114 \\ 
JVRC-1 & 90.78 & 85.74 & 0.071 & 0.103 & 0.095 \\ 
G1 & 97.83 & 94.08 & 0.025 & 0.088 & 0.102 \\

\end{tabular}
}
\caption{Quantitative evaluation of our universal agent with domain randomization for a large variety of human motions and five different bipedal robots. }
\label{tab:universal_quantiative}
\end{table*}

\begin{figure*}
    \centering
    \resizebox{0.98\textwidth}{!}{%
    \begin{tikzpicture}
        \begin{groupplot}[
            group style={
                group size=4 by 1,
                vertical sep=1.5cm,
                horizontal sep=1.5cm,
                x descriptions at=edge bottom,
            },
            xlabel={Training Steps},
            grid=major,
            width=0.5\linewidth,
            height=7cm,
            xlabel style={font=\LARGE},
            ylabel style={font=\LARGE},
            title style={font=\LARGE},
            ticklabel style={font=\large},
            legend style={
                at={(2.2, -0.3)},  
                draw=none, 
                anchor=north,
                font=\LARGE,
                cells={anchor=west},
                legend columns =-1,
            },
        ]
        
        \nextgroupplot[title={Living Rate (\%) $\uparrow$}, ymin=0, ymax=1.1]
        \addplot [no markers, color={rgb,255:red,53;green,87;blue,27}, line width=2.5pt, dashed] table [x index=1, y index=2, col sep=comma] {plots/comparison.csv};
        \addplot [no markers, color={rgb,255:red,21;green,70;blue,118}, line width=2.5pt] table [x index=1, y index=3, col sep=comma] {plots/comparison.csv};
        \addplot [no markers, color=cyan, line width=2.5pt] table [x index=1, y index=4, col sep=comma] {plots/comparison.csv};
        \addplot [no markers, color={rgb,255:red,139;green,108;blue,19}, line width=2.5pt] table [x index=1, y index=5, col sep=comma] {plots/comparison.csv};
        \addplot [no markers, color={rgb,255:red,173;green,38;blue,32}, line width=2.5pt] table [x index=1, y index=6, col sep=comma] {plots/comparison.csv};
        \legend{
                ImitationNet \cite{imitationnet},
                Ours (w/ constraints),
                Ours (w/o constraints),
                DeepMimic \cite{deepmimic},
                AMP \cite{amp}
            }

        \nextgroupplot[title={Success Rate (\%) $\uparrow$}, ymin=0, ymax=1.1]
        \addplot [no markers, color={rgb,255:red,21;green,70;blue,118}, line width=2.5pt] table [x index=1, y index=7, col sep=comma] {plots/comparison.csv};
        \addplot [no markers, color=cyan, line width=2.5pt] table [x index=1, y index=8, col sep=comma] {plots/comparison.csv};
        \addplot [no markers, color={rgb,255:red,139;green,108;blue,19}, line width=2.5pt] table [x index=1, y index=9, col sep=comma] {plots/comparison.csv};
        \addplot [no markers, color={rgb,255:red,173;green,38;blue,32}, line width=2.5pt] table [x index=1, y index=10, col sep=comma] {plots/comparison.csv};
        
        \nextgroupplot[title={Style Distance (m) $\downarrow$}, ymajorgrids]
        \addplot [no markers, color={rgb,255:red,21;green,70;blue,118}, line width=2.5pt] table [x index=1, y index=11, col sep=comma] {plots/comparison.csv};
        \addplot [no markers, color=cyan, line width=2.5pt] table [x index=1, y index=12, col sep=comma] {plots/comparison.csv};
        \addplot [no markers, color={rgb,255:red,139;green,108;blue,19}, line width=2.5pt] table [x index=1, y index=13, col sep=comma] {plots/comparison.csv};
        \addplot [no markers, color={rgb,255:red,173;green,38;blue,32}, line width=2.5pt] table [x index=1, y index=14, col sep=comma] {plots/comparison.csv};
        \nextgroupplot[title={Root Distance (m) $\downarrow$}, ymajorgrids]
        \addplot [no markers, color={rgb,255:red,21;green,70;blue,118}, line width=2.5pt] table [x index=1, y index=15, col sep=comma] {plots/comparison.csv};
        \addplot [no markers, color=cyan, line width=2.5pt] table [x index=1, y index=16, col sep=comma] {plots/comparison.csv};
        \addplot [no markers, color={rgb,255:red,139;green,108;blue,19}, line width=2.5pt] table [x index=1, y index=17, col sep=comma] {plots/comparison.csv};
        \addplot [no markers, color={rgb,255:red,173;green,38;blue,32}, line width=2.5pt] table [x index=1, y index=18, col sep=comma] {plots/comparison.csv};
        \end{groupplot}
                
    \end{tikzpicture}
    }
    \caption{\textbf{Quantitative evaluation} of DeepMimic \cite{deepmimic} and AMP \cite{amp} versus our proposed approach, with and without constraints.}
    \label{fig:variable_comparison}
\end{figure*}

\subsubsection{Comparison with baselines}\label{sec:comparison}
To address the lack of a standardized benchmark for human imitation in robots, we trained DeepMimic \cite{deepmimic} and AMP \cite{amp} using reference motions from ImitationNet \cite{imitationnet} on the BRUCE robot. Given the single-policy-per-motion design in DeepMimic, we selected walking and dancing motions for evaluation. Table \ref{tab:various_robots_quantitative} and Figure \ref{fig:variable_comparison} present the results. As noted in \cite{deepmimic}, DeepMimic is highly sensitive to reward weights. In contrast, our I-CTRL model exhibits robustness across robot types and motion variations without requiring additional tuning.

For DeepMimic, we perform extensive hyperparameter tuning to optimize walking motion performance and apply the same configuration to dancing. Despite this effort, DeepMimic underperformed relative to I-CTRL.

\begin{figure}[]
    \centering
    \includegraphics[width=0.98\linewidth]{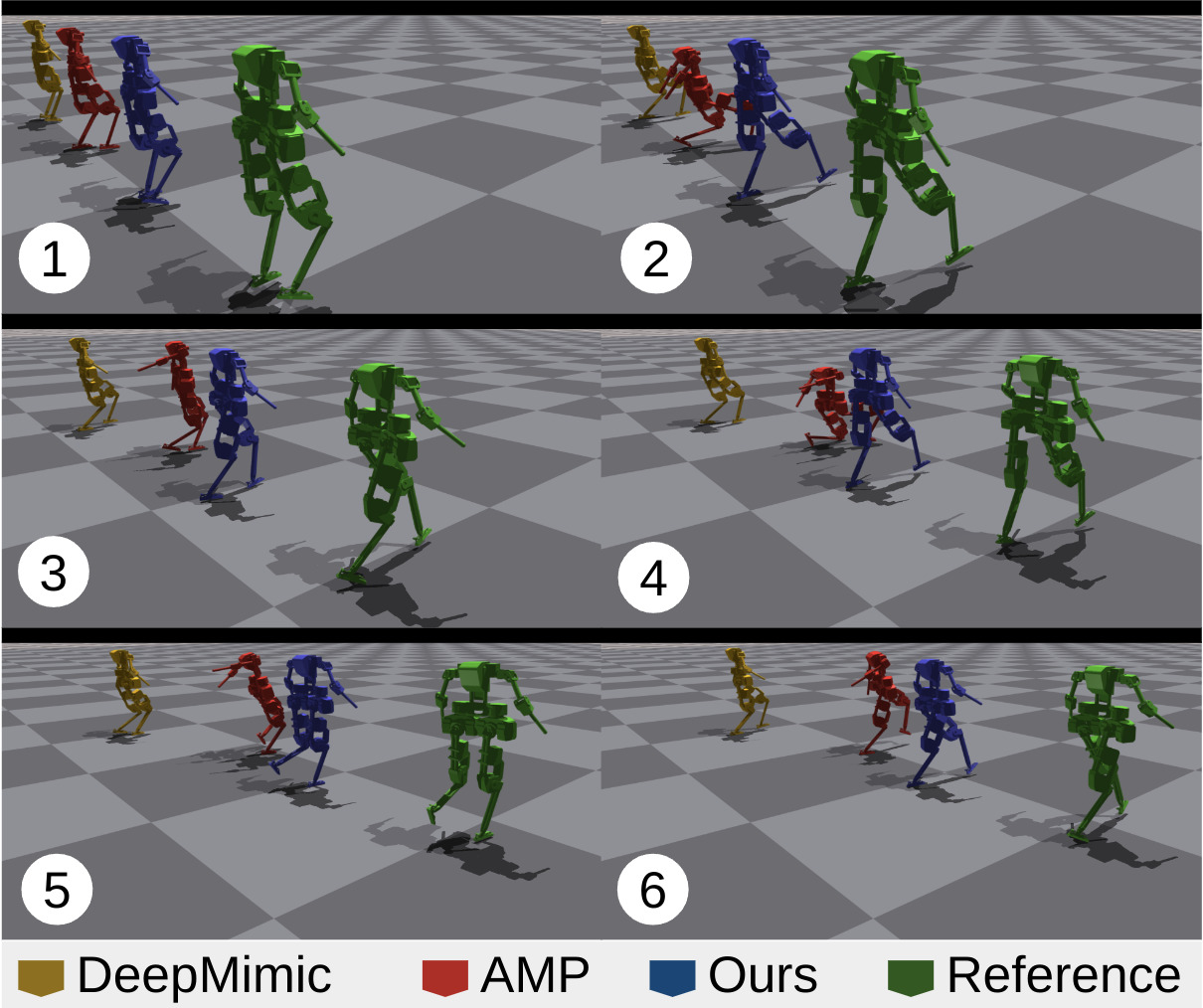}
    \caption{\textbf{Comparison of the performance of DeepMimic \cite{deepmimic} and AMP \cite{amp} versus our approach.} Here, the reference refers to the retargeted robot motion from a real human walking using ImitationNet \cite{imitationnet}. The reference robot (green) ignores the physics laws for the visualization purpose.}
    \label{fig:comparison_figure}
\end{figure}

We also compared I-CTRL with AMP \cite{amp}, which uses a discriminator network to reward indistinguishable motions from reference. As shown in Table \ref{tab:various_robots_quantitative}, I-CTRL outperforms AMP for both walking and dancing motions. While AMP and DeepMimic were initially trained on highly accurate reference motions, our approach uses outputs from a motion retargeting algorithm like ImitationNet, which introduces noise. This discrepancy between noisy references and physics-based motion is easily detected by AMP's discriminator. Indicators such as foot penetration or lack of ground contact in our reference motions may simplify the discriminator's task. Figure \ref{fig:comparison_figure} illustrates this comparison for the straight-line walking motion.

\begin{table}[]
\centering
\resizebox{0.99\linewidth}{!}{%
\begin{tabular}{@{}lcccc@{}}
 \textbf{Policy} & \textbf{Success (\%)}  & \textbf{Style (m)}  &
\textbf{Root Pos. (m)} &
\textbf{Root Rot. (\degree)}  \\ \midrule
\multicolumn{5}{c}{\textbf{Walking}}  \\ \midrule
DeepMimic & 72.07 & 0.031 & 0.136 & \textbf{0.557} \\
AMP   & 59.62 & 0.094 & 0.169 & 9.710 \\ 
Ours &\textbf{100.00} & \textbf{0.010} & \textbf{0.036} & 4.263\\ \midrule
 \multicolumn{5}{c}{\textbf{Dancing}}  \\ \midrule
DeepMimic & 97.94 & 0.050 &  0.112 & 2.510 \\
AMP  & 98.92 & 0.089 & 0.098 & 19.54  \\ 
Ours  & \textbf{100.00} &\textbf{ 0.049} & \textbf{0.039} & \textbf{1.036} \\ \midrule
\end{tabular}
}
\caption{Quantitative evaluation of our Bruce robots in the single policy training for a walking and dancing motion. Here, bold denotes the best result per metric and motion. Note that this table simply validates our constrained technique versus prior baselines \cite{deepmimic, amp}, and we do not use privileged information for the critic module, APS, and domain randomization.}
\label{tab:various_robots_quantitative}
\end{table}

{\subsubsection{Robustness to noisy reference} We address the question of the effectiveness of I-CTRL under low-quality reference motions.  As shown in Sec. \ref{sec:comparison}, DeepMimic and AMP require a high-precision MoCaP reference motion to achieve high-quality imitation, and underperform when the reference motion is obtained from a motion retargeting algorithm, such as ImitationNet \cite{imitationnet}. In order to evaluate the robustness of I-CTRL under poor reference motions, we artificially add Gaussian noises at different scales to the reference motion from ImitationNet during the evaluation. Precisely, $\mathbf{\tilde{q}}_{1:T}^{ref} = \mathbf{q}_{1:T}^{ref}  + G$ where $G \sim \mathcal{N}(0,\,\sigma^{2})$. 
We report our quantitative results in Table \ref{tab:noise} for different standard deviations ($\sigma$). As we expected, as the noise level increases, it also augments the uncertainty of I-CTRL in the task performance. However, the BRUCE robot is still able to successfully follow the walking trajectory in 73.1\% of the cases when the body has been affected with an additional Gaussian noise of $\sigma=5\degree$ on top of the errors of the imitation algorithm.
}
 
\begin{table}[]
\centering

\resizebox{0.48\textwidth}{!}{%
    \begin{tabular}{l|c|c|c}
    Noise - $\sigma$ (\degree) & 0 &  5 & 10\\ \midrule
     {Living Rate (\%) $\uparrow$} &  100 & 98.0 & 86.5\\
      {Success Rate (\%) $\uparrow$} & 91.1 & 73.1 &  60.6 \\
      {Root Pos. Dist. (m) $\downarrow$} & $0.10 \pm 0.06$  & $0.13 \pm 0.10$  & $0.15 \pm 0.11$  \\
      {Root Rot. Dist. (\degree) $\downarrow$} & $0.90 \pm 0.42$  & $1.50 \pm 2.22$ & $3.01 \pm 4.80$  \\
    \end{tabular}
}
\caption{ Quantitative evaluation of the robustness of our model in following a walking forward motion given different noise rates in the reference style motion for the BRUCE robot.}
\label{tab:noise}
\end{table}

\subsection{Future work} 
Our I-CTRL framework enables robots to imitate diverse human motions while preserving high visual resemblance. However, our system only accounts for locomotion behaviors without interactions with complex environments. Investigating a humanoid bipedal agent that can learn daily loco-manipulation skills from humans, such as carrying a box or opening a door, will be considered in the future.

\section{Conclusion} 
\label{sec:conclusion}

\noindent In this paper, we introduced a novel physics-based motion retargeting framework through bounded residual reinforcement learning to imitate human motion on various legged humanoid robots, named I-CTRL. We consider the imitation task as a two-stage problem: first, we retarget human motions to robots and then refine the infeasible motions using I-CTRL. Therefore, we constrain the exploration in joint space for the RL agent to a smaller $\delta$-ball subspace, which highly improves motion resemblance (reducing 80\% the style error) and success rate (+5.62\%). Additionally, we focus on a versatile imitation system that generalizes to different bipedal robots without reward function fine-tuning, from small-sized BRUCE to full-sized H1, and from lightweight robots like G1 and JVRC-1 to the heavyweight ATLAS. Under consistent settings, our method enables these robots to imitate, with a single policy trained in only 15 hours, large-scale complex dynamic motions such as walking in circles, fast running, jumping, or miscellaneous dances, among others. Thanks to I-CTRL, robots can imitate an average of 82.94\% of the motions while achieving high visual similarity and preserving root trajectories. For example, the BRUCE robot follows the reference motions with its joint and root positions deviating by less than 1.9 cm and 9 cm, respectively, from the reference motion. This demonstrates that the generated physics-based motions maintain visual similarity and accurately replicate root trajectories. These behaviors are validated through various quantitative and qualitative experiments across different motions and robots under domain randomization to facilitate sim-to-real transfer.

\section*{Acknowledgements} 
\noindent This work is funded by Marie Sklodowska-Curie Action Horizon 2020 (Grant agreement No. 955778) for project 'Personalized Robotics as Service Oriented Applications' (PERSEO).

\bibliographystyle{plainnat}
\bibliography{references}


\begin{IEEEbiography}[{\includegraphics[width=1in,height=1.25in,clip,keepaspectratio]{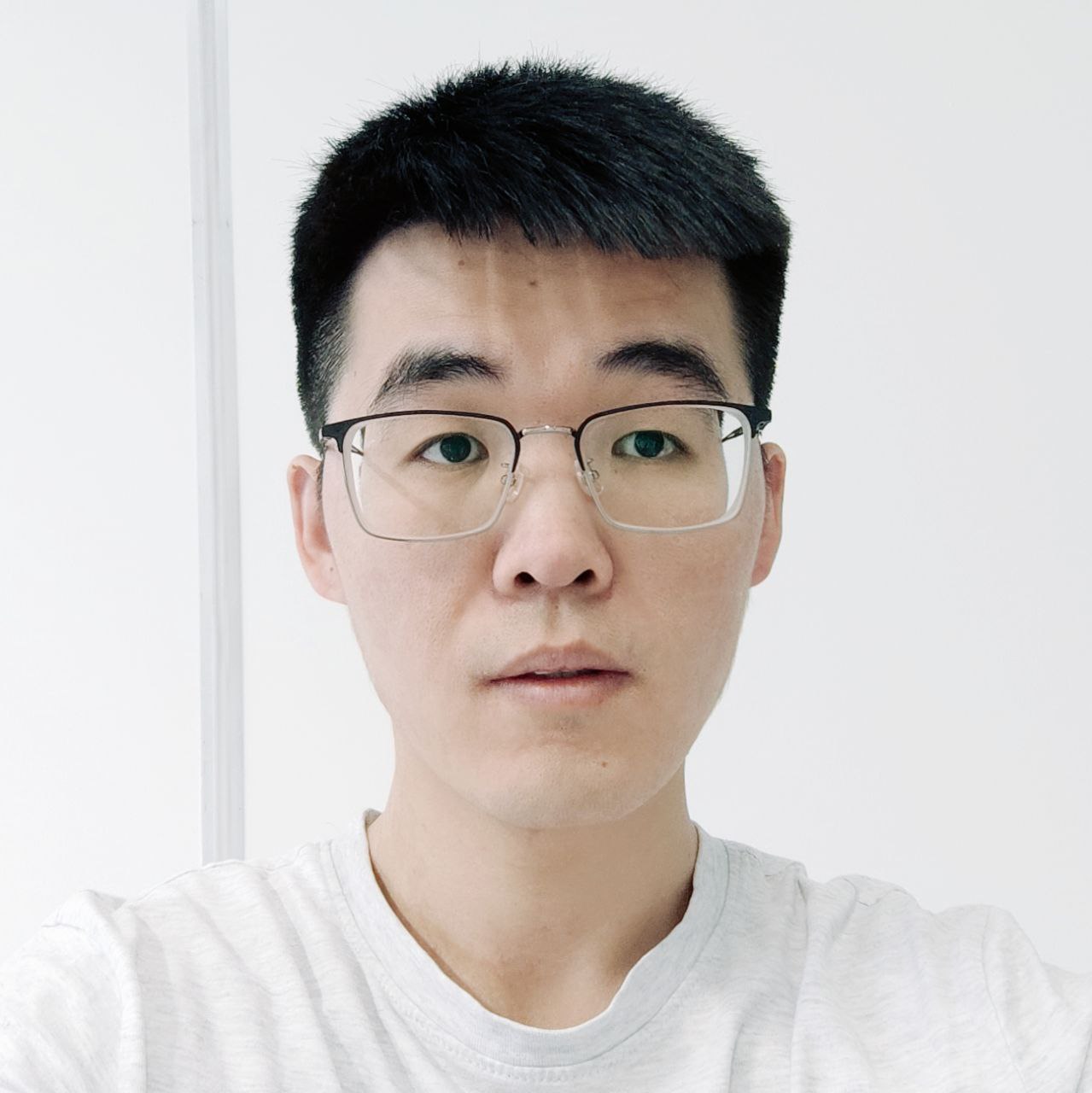}}]
{Yashuai Yan} received his B.Sc. degree in Computer Science from Technical University Braunschweig, Braunschweig, Germany, and his M.Sc. in Data Analytics and Engineering from the Technical University of Munich (TUM), Munich, Germany, 2020 and 2023, respectively. Afterward, he joined Autonomous Systems at the Technische Universität Wien (TU Wien) to work in the field of robotics. His research focuses on robot learning from human demonstrations, as well as AI-embedded robotic systems.
\end{IEEEbiography}

\begin{IEEEbiography}
[{\includegraphics[width=1in,height=1.25in,clip,keepaspectratio]{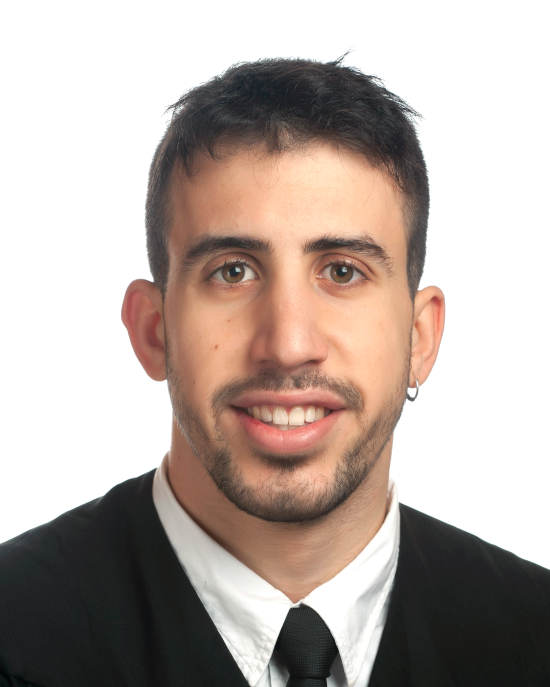}}]
{Esteve Valls Mascar\'o} received a B.S. and M.S degree in telecommunications engineering from Universitat Politecnica de Catalunya (UPC) in 2020 and 2021, respectively, while he worked in several companies as a computer vision engineer. He joined the Human-Centered Assistive Robotics Group at the Technical University of Munich (TUM), Munich, Germany in 2021, and later the Autonomous Systems Lab at Technische Universität Wien (TU Wien), Vienna, Austria, to work towards his Ph.D. degree. His research is focused on understanding the human intention through AI for a better human-robot interaction.
\end{IEEEbiography}

\begin{IEEEbiography}[{\includegraphics[width=1in,height=1.25in,clip,keepaspectratio]{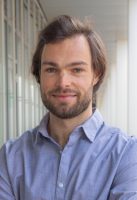}}]
{Tobias Egle} received his M.Sc. in Mechatronics and Robotics from the Technical University Munich (TUM) Germany in 2022. He joined the Automation and Control Institute (ACIN) at the TU Wien as a researcher in 2022. His research interests include legged locomotion, model- and data-based motion planning, humanoid robots, and whole-body control.
\end{IEEEbiography}

\begin{IEEEbiography}[{\includegraphics[width=1in,height=1.25in,clip,keepaspectratio]{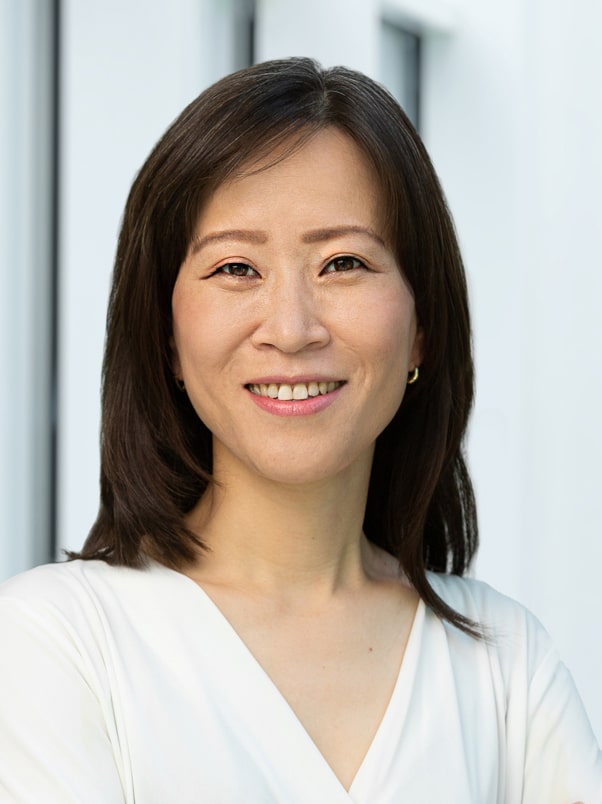}}]{Dongheui Lee} is a Full Professor of Autonomous Systems at TU Wien since 2022. She is also leading the Human-Centered Assistive Robotics group at the German Aerospace Center (DLR), Institute of Robotics and Mechatronics, since 2017. Her research interests include human motion understanding, human-robot interaction, machine learning in robotics, and assistive robotics. Prior to her appointment at TU Wien, she was an Assistant Professor and Associate Professor at the Technical University of Munich (TUM), a Project Assistant Professor at the University of Tokyo, and a research scientist at the Korea Institute of Science and Technology (KIST). She obtained a PhD degree from the Department of Mechano-Informatics, University of Tokyo in Japan. She was awarded a Carl von Linde Fellowship at the TUM Institute for Advanced Study and a Helmholtz professorship prize. She has served as Senior Editor and a founding member of IEEE Robotics and Automation Letters (RA-L) and Associate Editor for the IEEE Transactions on Robotics.

\end{IEEEbiography}

\end{document}